\documentclass{article}
\usepackage{url}
\usepackage[pdftex]{graphicx}
\usepackage{amsmath}
\usepackage{booktabs}
\usepackage{amsfonts}
\usepackage{algorithm}
\usepackage{algpseudocode} 
\usepackage{float}
\usepackage{natbib}
\usepackage{multirow}
\usepackage{tcolorbox}
\usepackage{authblk}  
\usepackage{booktabs}      
\usepackage{adjustbox}     
\usepackage{caption}       
\usepackage{tabularx}      

\title{Towards Agents That Know When They Don't Know: Uncertainty as a Control Signal for Structured Reasoning}
\author[1]{Josefa Lia Stoisser\thanks{Equal contribution.}}
\author[1]{Marc Boubnovski Martell\textsuperscript{*}}
\author[1]{Lawrence Phillips}
\author[1]{Gianluca Mazzoni}
\author[1]{Lea Mørch Harder}
\author[2]{Philip Torr}
\author[1]{Jesper Ferkinghoff-Borg}
\author[1]{Kaspar Märtens\thanks{Equal contribution.}}
\author[1]{Julien Fauqueur\textsuperscript{$\dagger$}}

\affil[1]{Novo Nordisk}
\affil[2]{University of Oxford}

\date{September 2025} 

\begin{document}

\maketitle

\begin{abstract}
Large language model (LLM) agents are increasingly deployed in structured biomedical data environments, yet they often produce fluent but overconfident outputs when reasoning over complex multi-table data. We introduce an uncertainty-aware agent for query-conditioned multi-table summarization that leverages two complementary signals: (i) retrieval uncertainty—entropy over multiple table-selection rollouts—and (ii) summary uncertainty—combining self-consistency and perplexity. Summary uncertainty is incorporated into reinforcement learning (RL) with Group Relative Policy Optimization (GRPO), while both retrieval and summary uncertainty guide inference-time filtering and support the construction of higher-quality synthetic datasets.

On multi-omics benchmarks, our approach improves factuality and calibration, just less than tripling correct and useful claims per summary (3.0→8.4 internal; 3.6→9.9 ulti-omics) and substantially improving downstream survival prediction (C-index 0.32→0.63). These results demonstrate that uncertainty can serve as a control signal—enabling agents to abstain, communicate confidence, and become more reliable tools for complex structured-data environments.
\end{abstract}

\section{Introduction}

Imagine a biomedical researcher querying a large multi-omics database to identify candidate biomarkers for survival outcomes \cite{jin2024genegpt}. A standard LLM-based agent may confidently produce a fluent statement such as “gene X is strongly associated with survival in patients”—even when the underlying tables contain contradictory or insufficient evidence. To the end user, this confident but unqualified claim is indistinguishable from a reliable finding \cite{omar2025benchmarking,martell2025scalable}. By contrast, an uncertainty-aware agent could detect the inconsistency, flag its own low confidence, or abstain altogether\cite{zhao-etal-2025-uncertainty,zhao2024saup,hu2024uncertainty, han2024towards}. This ability to communicate not only what is said but also how certain it is transforms raw text generation into actionable, trustworthy scientific insight \cite{bolton2024rambla, omar2025benchmarking,hakim2024need}.

Most modern scientific knowledge is encoded not in natural language but in high-dimensional tables such as genomic assays, proteomic screens, and electronic health records \citep{gtex2020gtex, bycroft2018uk, kang2022roadmap, probst2020visualization}. These resources contain invaluable information that could accelerate biomedical discovery, yet they remain largely inaccessible to non-specialists. Extracting meaningful insights from such data, i.e. generating summaries, demands not only computational power but also the ability to translate complex numerical signals into coherent narratives—an area where LLMs are uniquely positioned to contribute \citep{li2025conceptual, yu2025tablerag}. The novelty of our work lies in using uncertainty-aware signals to both calibrate agents and filter summary outputs, enabling their use as synthetic data \citep{lee2025synthetic}. This approach enhances the quality of training corpora, ultimately enabling more robust and reliable downstream decision-making.

Recent work has begun adapting LLMs for tabular summarization and reasoning. Query-focused methods such as QTSumm \citep{zhao2023qtsumm} generate targeted textual insights from structured inputs, while StructText \citep{kashyap2025structtext} and eC-Tab2Text \citep{guanilo2025ec} introduce synthetic benchmarks across scientific and e-commerce domains. Evaluation frameworks such as FineSurE \citep{song2024finesure} and multi-agent debate approaches \citep{estornell2024multi} reveal the challenges in measuring faithfulness and coverage in generated summaries, highlighting the limitations of current single-pass generation methods \cite{sui2025chain}.

An emerging paradigm involves designing table agents—LLM-driven systems that integrate structured querying, strategic planning, and external tool use \cite{bendinelli2025exploring,mathur2024matsa, stoisser2025query}. For example, \cite{lu2025large} outline design principles for real-world table agents capable of combining SQL execution with reasoning chains, while  demonstrate multi-agent orchestration for multi-document reasoning tasks \cite{sui2025chain}. Beyond summarization, frameworks such as MAG-V \citep{sengupta2024mag} exemplify iterative generation and verification of synthetic data, illustrating a blueprint for refinement over one-shot output.

However, these promising approaches share a critical blind spot: uncertainty. LLMs are known to produce fluent yet unfaithful outputs \citep{xu2024hallucination}, a problem exacerbated when summarizing high-dimensional data \citep{fang2024large, wu2025tabular}. We conceptualize uncertainty quantification (UQ) as a form of agent–environment interaction \citep{han2024towards}, where the focus is not only on data quality but also on the agent’s confidence and reliability in navigating complex tables. Recent efforts in UQ range from confidence–consistency scoring methods such as CoCoA \citep{vashurin2025uncertainty} to head-based uncertainty prediction (RAUQ, UQLM \citep{vazhentsev2025uncertainty, bouchard2025uqlm}). Other works explore faithfulness-aware UQ in retrieval-augmented generation \citep{fadeeva2025faithfulness} and structured tasks such as text-to-SQL \citep{somov2025confidence}, underscoring the necessity of calibration for trustworthy table understanding.

In this paper, we propose an uncertainty-aware LLM agent for summarizing high-dimensional tabular data. Our agent generates candidate summaries from multi-omics datasets, quantifies its own uncertainty, and filters outputs with high uncertainty. We evaluate the approach on biomedical multi-omics tasks, where multiple valid summaries exist—highlighting the critical role of calibration beyond mere coverage.

Our contributions are threefold:
\begin{enumerate}
    \item \textbf{Uncertainty as control:} We introduce the first LLM agent framework where uncertainty is not just monitored but directly used as a reward signal during training, and as an abstention/filtering signal at inference, moving beyond post-hoc diagnostics.     
    \item \textbf{Robustness in structured environments:} On biomedical multi-omics tasks, uncertainty-aware agents achieve higher factuality, calibration, and downstream utility, with methods applicable to any multi-table setting.  
    
    \item \textbf{Uncertainty as data-quality signal: }We show that filtering high-uncertainty samples improves tabular text dataset quality, providing a practical tool for curating reliable corpora. 
\end{enumerate}

\begin{figure}[t]
    \centering
    \includegraphics[scale=0.34]{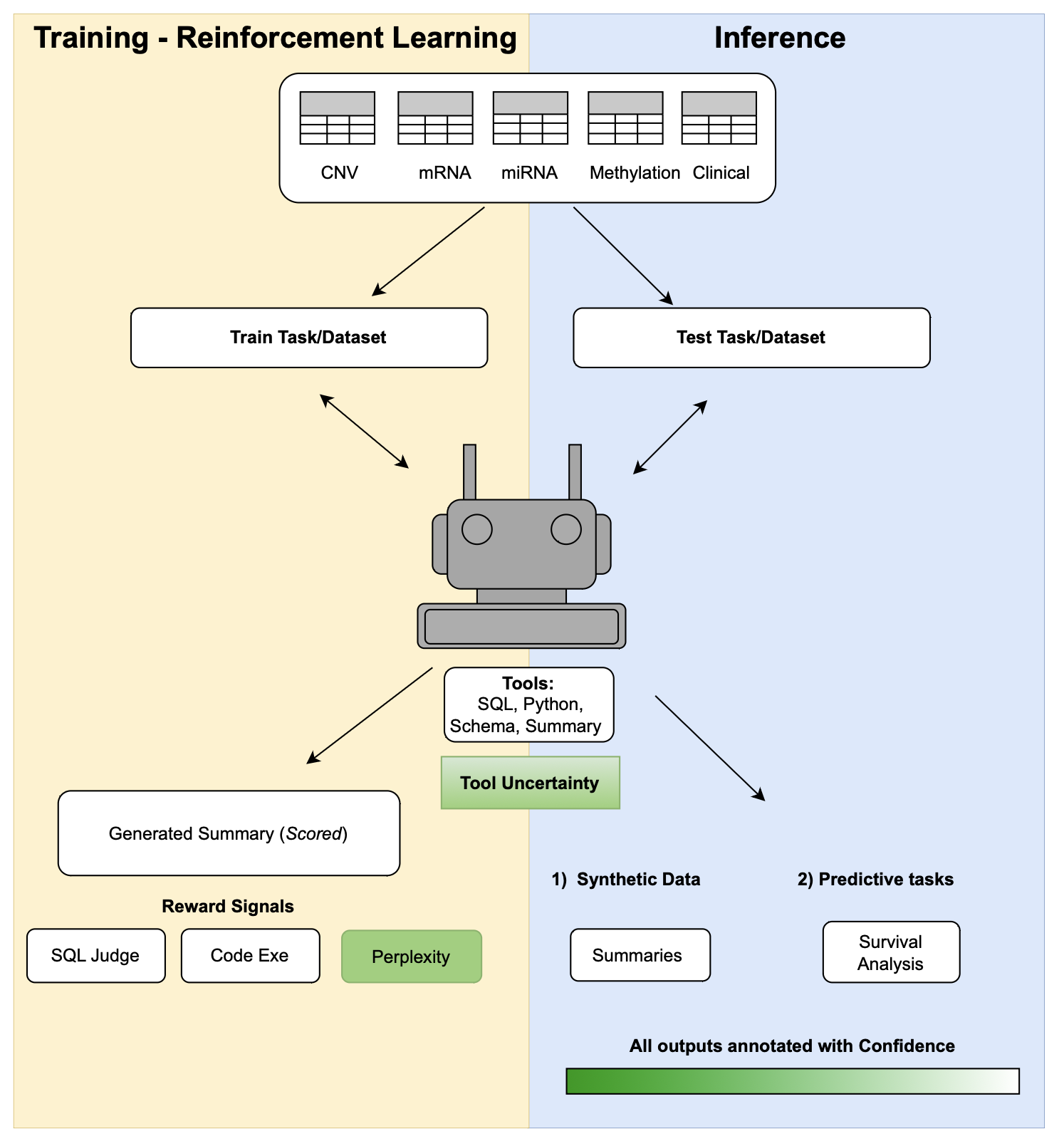}
    \caption{
    \textbf{The Uncertainty-Aware Agent Framework.}
This diagram shows the two phases of our agent: \textbf{(a) training with reinforcement learning}, and \textbf{(b) inference}. In training, the agent's policy is refined using a reward signal informed by summary uncertainty (perplexity). During inference, multiple rollouts generate candidate summaries, which are then filtered based on a combined score of retrieval and summary uncertainty, leading to more reliable outputs.
    }
    \label{fig:main}
\end{figure}

\section{Background}
\label{gen_inst}
\subsection{Interactive agent frameworks for structured reasoning}

Early table summarization methods primarily relied on rule-based or statistical approaches, producing template-based outputs and lacking explicit uncertainty modeling. Recent advances employ neural and LLM-based methods that shift from static, single-pass generation to interactive reasoning over structured environments. For example, LLM agents can now issue SQL queries or dataframe operations \cite{stoisser2025sparks, stoisser2025struct}, dynamically retrieving evidence before forming summaries. Surveys of table agents \cite{tian2025toward}  highlight how symbolic querying and neural reasoning can be combined to support exploratory analysis and hypothesis generation. More recently, multi-agent frameworks such as MAG-V (generator–verifier) \citep{sengupta2024mag} and Multi2 (scalable multi-document reasoning)\citep{cao2025multi2} demonstrate how dividing labor among specialized agents can improve reliability and scalability. These works suggest that interactive, tool-augmented agents are a promising direction for table understanding.

\subsection{Uncertainty quantification in LLMs}
Despite progress in interactivity, most agents remain prone to overconfidence and unfaithful outputs. Traditional metrics such as BLEU or ROUGE fail to capture factual reliability in structured domains. This has led to the development of uncertainty quantification approaches and libraries such as CoCoA \citep{vashurin2025uncertainty} and LM-Polygraph \citep{fadeeva2023lm}, which use probabilistic confidence and/or semantic self-consistency to detect hallucinations. In structured tasks like text-to-SQL \citep{shorinwa2025survey}, confidence estimation has been shown to prevent execution errors by flagging low-confidence predictions \cite{maleki2025confidence}. Similarly, in retrieval-augmented generation, uncertainty-aware thresholds can trigger additional retrieval or abstention \citep{fadeeva2025faithfulness, soudani-etal-2025-uncertainty}. However, most of these methods treat uncertainty as a post-hoc diagnostic \cite{hao2025uncertainty}. They are not integrated into the agent’s decision-making process during interaction with tables, limiting their effectiveness in dynamic environments\cite{hu2025agentgen}.

\subsection{Toward self-assessment in scalable agents}
A growing body of work suggests that scalable and trustworthy agents must go beyond post-hoc uncertainty estimation toward learned self-assessment \cite{han2024towards, guan2024richelieu, renze2024self}. Active learning studies \citep{melo2024deep,ye2025uncertainty} show that focusing on uncertain cases improves efficiency, while debate-style multi-agent systems \citep{yin2025enhancing}  demonstrate how structured disagreement enhances reliability. Recent explorations of self-reflection in LLMs indicate that agents can improve reasoning by monitoring their own confidence \cite{liu2025truth}. Yet, existing work has not combined these insights into a framework where uncertainty directly controls both training optimization and inference-time behavior \cite{cui2025entropy, zhang2025no}. Structured domains such as databases, where repeated querying and summarization are natural, provide fertile ground for such uncertainty-aware self-assessment. This paper builds on these insights by proposing a framework in which retrieval stability and output consistency are treated as first-class control signals, enabling LLM agents to produce more reliable and trustworthy multi-table summaries.

\section{Methods}

We cast query-conditioned multi-table summarization as an episodic agent problem
and make \emph{uncertainty} a control signal: We (i) measure retrieval instability
and output inconsistency, (ii) shape training rewards with those signals, and
(iii) apply them during inference to filter summaries and enrich them with a quality signal.


\subsection{Problem formulation}
Let $\mathcal{D}$ be a structured database and $q$ a natural-language task. A policy
$\pi_\theta$ interacts with $\mathcal{D}$ via tools and emits a summary $s$ that encapsulates the information in the database relevant to the given task:
\[
(q,\mathcal{D})\xrightarrow{\ \pi_\theta\ } s.
\]

\subsection{Environment and Episode Setup}
\label{sec:agent_env}
Each episode takes place in an environment consisting of:  
(i) a structured database $\mathcal{D}$ containing tables, columns, and descriptions, and  
(ii) a task $q$.  
At timestep $t$, the state $x_t$ includes the task $q$, the schema snapshot of $\mathcal{D}$, and the history of previous actions and results. The agent selects actions $a_t \sim \pi_\theta(a_t \mid x_t)$, which the environment executes deterministically. Available actions are:
\begin{itemize}
   \item \textbf{SQLExecutor(query)} – Executes a SQL query to retrieve or join rows across tables in $\mathcal{D}$.
   \item \textbf{Schema(table)} – Returns the structure, column names, and types of a specified table.
   \item \textbf{PythonTool(code)} – Runs Python code to process query results or perform computations when SQL is insufficient.
   \item \textbf{CommitSummary(summary)} – Terminates the episode and outputs a final summary $s$.
\end{itemize}
\paragraph{Episode flow.}  
An episode thus consists of a query, a sequence of tool calls, and a terminating summary. Formally, invoking \texttt{CommitSummary} yields a trajectory
\[
\tau = \big((x_0,a_0), (x_1,a_1), \dots, (x_T,a_T)\big)
\]
and a final output $s$. During \textit{training}, trajectories are scored under GRPO with rewards combining (i) code correctness, (ii) exploration coverage of $\mathcal{D}$, and (iii) confidence in the summary (measured by perplexity). During \textit{inference}, we sample multiple trajectories per query. Uncertainty is estimated via retrieval entropy and CoCoA; if uncertainty is high, the agent abstains. Otherwise, the lowest-perplexity summary is returned, accompanied by confidence scores. Full algorithmic details are in Algorithm \ref{alg:epi} Appendix~\ref{app:details}.

\subsection{Uncertainty Signals}

\paragraph{Summary uncertainty (training: Perplexity, inference: CoCoA).}
We adopt perplexity-based {CoCoA} from \cite{vashurin2025uncertainty}, which unifies two signals: token-level confidence (perplexity) and semantic consistency across samples. The resulting Minimum-Bayes-Risk-derived score $u_{\text{CoCoA}}$ aligns more strongly with true error rates than either component alone. At inference, we sample \( K \) candidate summaries and compute CoCoA to accept or abstain. By construction, CoCoA already integrates perplexity, so no separate perplexity term is calculated at inference; during training, we use perplexity $u_{\text{Perp}}$ alone as a cheaper proxy. Full details are described in Appendix \ref{app:details}.

\underline{Example (CoCoA).}
For a query on ``biomarkers associated with survival in cancer patients'', one episode yields ``The upregulation of genes X, Y, and Z is associated with a significant decrease in predicted survival time for patients with aggressive cancer types,'' while another outputs ``The expression levels of genes X, Y, and Z show no correlation with survival outcomes across the patient cohort.'' Low cross-sample consistency raises the CoCoA score, signaling semantic inconsistency and triggering abstention despite both trajectories being individually plausible.

\paragraph{Retrieval uncertainty (inference-only).} High-dimensional databases pose challenges in table selection; we address this by quantifying retrieval uncertainty.
For a fixed task $q$, run $K$ retrieval episodes. Let $R^{(k)}$ be the set of
tables touched in episode $k$, and define the candidate set
$C=\bigcup_{k=1}^K R^{(k)}$. The empirical selection frequency for $t\in C$ is
$\hat p_t=\tfrac{1}{K}\sum_{k=1}^K \mathbf{1}[t\in R^{(k)}]$. We compute
normalized binary entropy
$H(t)=-\tfrac{\hat p_t\log \hat p_t+(1-\hat p_t)\log(1-\hat p_t)}{\log 2}$ and
aggregate
\begin{equation}
u_{\text{ret}}(q)=\tfrac{1}{|C|}\sum_{t\in C} H(t).
\label{eq:uret}
\end{equation}
High $u_{\text{ret}}$ indicates inconsistent evidence acquisition. We compute
$u_{\text{ret}}$ during inference but omit it as a training reward due to the high computational cost of sampling.

\underline{Example (Retrieval Uncertainty).}
For query ``biomarkers associated with survival in cancer~X'', the agent first
invokes \texttt{SQLExecutor} to retrieve candidate gene--expression tables. It
then issues a second targeted SQL to join clinical survival labels. If repeated
episodes select different tables, retrieval uncertainty $u_{\text{ret}}$ is high,
indicating unstable evidence and triggering abstention at inference.

\subsection{Training rewards}

We use three terminal reward components: (i) \emph{Code execution} which rewards the agent for correctly executing SQL queries and Python code, teaching it to effectively navigate the environment;  (ii) an \emph{LLM-judge} score, which promotes broad, grounded factual coverage, encouraging exploration of the dataset environment for information; and (iii) \emph{summary confidence}, which favors low-uncertainty summaries, promoting the exploitation of existing knowledge. The reward is a weighted sum
$R(\tau)=\alpha_{\text{code}}R_{\text{code}}+\alpha_{\text{judge}}R_{\text{judge}}+\alpha_{\text{conf}}R_{\text{conf}}$. Formulas and weights are given in Appendix \ref{app:details}.

\paragraph{Schedules.}
To balance exploration ($R_{\text{Judge}}$) and exploitation ($R_{\text{conf}}$) over the 100 training steps $t$, we make $\alpha_{\text{conf}}$ depend on $t$ and introduce reward schedules. The \textit{Baseline} schedule ($R\equiv R_{base}$) applies fixed weights throughout training but risks harming early exploration of the dataset. \textit{Two-Phase} ($R\equiv R_{phase}$) prioritizes exploration in early steps and adds exploitation midway through. \textit{Stepwise Addition} ($R\equiv R_{step}$) periodically boosts $R_{\text{conf}}$ at regular intervals, while retaining exploration focus. \textit{Adaptive Exploitation} ($R\equiv R_{adapt}$) dynamically adjusts $\alpha_{\text{conf}}$ based on intermediate $R_{\text{Judge}}$ performance, integrating continuous exploitation that gradually tapers off as summaries stabilize. See Table~\ref{tab:reward-schedules} in Appendix \ref{app:details} for details.

\subsection{Optimization with GRPO}
We train with \emph{Group Relative Policy Optimization} (GRPO), a PPO-style
objective with a KL penalty to a reference policy $\pi_{ref}$, effective for reasoning LLMs \citep{shao2024deepseekmath,guo2025deepseek,liu2025understanding,singh2025agentic}. With ratio $r_\theta(\tau)=\pi_\theta(\tau)/\pi_{\text{old}}(\tau)$, we maximize
\begin{equation}
\mathcal{L}(\theta)=\mathbb{E}_\tau\Big[\min\big(r_\theta A,\ \text{clip}(r_\theta,1-\epsilon,1+\epsilon)\,A\big)\Big]
-\beta\,D_{\mathrm{KL}}(\pi_\theta\Vert\pi_{\text{ref}}),
\label{eq:grpo}
\end{equation}
where $A\equiv A(\tau)$ is the advatage of trajectory $\tau$, derived from the reward $R(\tau)$.

\subsection{Inference: Post-output filtering}
\label{sec:inference}

At inference we sample $K$ trajectories, compute $u_{\text{ret}}$ and
$u_{\text{CoCoA}}$, and apply a conservative rule: abstain if the sum
exceeds a tuned threshold $2 \kappa$; otherwise emit the candidate with lowest
$u_{\text{Perp}}$ and use $u_{\text{ret}}$ and
$u_{\text{CoCoA}}$ as reliabilty scores. Threshold values are determined on a validation split through human inspection. Details are described in algorithm~\ref{alg:epi} in Appendix \ref{app:details}.

\section{Experiments}
\label{sec:experiments}


\subsection{Datasets}
We evaluate our approach on two multi-omics databases: one public benchmark and one internal, proprietary dataset. The MLOmics benchmark, which focuses on cancer research, has a flat structure with only 45 tables, and consists mostly of raw measurements, while the internal dataset features a tree-like schema with over 2{,}000 tables and includes aggregated summary statistics. This diversity allows us to assess whether our agent remains robust across (i) compact, unprocessed data scenarios, and (ii) highly structured, large-scale environments, as the schema is shown in Appendix \ref{app:datasest}.

For agents, evaluating across multiple environments is critical: policies often overfit to the dynamics of a single environment schema and fail to generalize when the relational structure or data granularity changes \cite{subbaswamy2021evaluating, jiang2023importance}. Recent work on environment generalization in RL \cite{gu2025robust, teoh2025generalization} shows that agents trained in one setting may exploit spurious regularities and collapse when exposed to even minor distributional shifts. In line with these findings, we deliberately test on both a compact raw benchmark and a large schema-rich dataset to probe whether our approach adapts robustly to environment variation.

\paragraph{MLOmics dataset.}
We also evaluate on the MLOmics benchmark \cite{yang2025mlomics}, an open cancer multi-omics dataset with 8,314 patient samples across 32 cancer types. It provides four modalities—mRNA, microRNA, DNA methylation, and copy number variation. We use the \emph{Top} feature version (ANOVA-selected subsets), which offers a standardized and reproducible public testbed complementing our internal dataset.
Details and visuals of the dataset schemas are available in Appendix~\ref{app:datasest}.

\paragraph{Internal multi-omics dataset.}
Our internal dataset stems from layered biomedical omics. While the contents are proprietary, it includes tens to thousands of tables across transcriptomics, proteomics, and metabolomics. The schema combines a tree-like hierarchy from root entities with a broad relational structure hinging on a central table—making it a compelling testbed for agent adaptability.

\subsection{Implementation Details}
The datasets are split into training and testing sets with a 70:30 ratio based on patient samples (Figure \ref{fig:main}), ensuring consistent representation of all tables. We define 100 summary tasks per dataset, validated by scientists (examples in Appendix \ref{app:task_templates}), evaluated by LLMs and domain experts, and designed to capture the most relevant information comprehensively. Of these, 80 tasks are used for training and 20 for evaluation. During inference, each task is answered five times, and we report the mean and standard deviation of the scores for robustness.

All experiments utilize the \texttt{ART} framework\footnote{\url{https://art.openpipe.ai/}}, with \texttt{Qwen2.5-14B-Instruct} employed as the policy backbone. Training is conducted on a single NVIDIA A100 GPU. Hyperparameters are discussed in greater detail in Appendix \ref{app:Hyper}. Each training episode allows for up to six tool calls prior to committing a summary. During inference, $K=5$ episodes are sampled per task to estimate retrieval and summary uncertainty.

\subsection{Metrics}
\label{sec:evaluation}
We evaluate the \textit{quality and uncertainty of summaries} and the \textit{reliability of uncertainty measures} as follows:

\textbf{Summary Quality.}
To quantify summary quality, we report three metrics: (Q1) the total number of claims, reflecting the summary’s richness in terms of content; (Q2) the ratio of correct claims, which measures factuality; and (Q3) the ratio of useful claims, which captures their relevance to the task.
We derive these metrics by decomposing the summary into claims that can be assessed by an LLM fact-checking judge, following evidence that LLM judges provide reliable and fine-grained evaluations \cite{xie2025empirical, zhou2025evaluating}. 
Specifically, given a summary \(s\), a task \(q\), and a database \(\mathcal{D}\), an o4 mini judge decomposes \(s\) into atomic claims, validates them against \(\mathcal{D}\) using a set of five task-specific workflows (designed in collaboration with domain experts), and assigns correctness and utility labels to each claim. 

\textbf{Uncertainty.}
To evaluate the model's confidence in its generated summaries, we compute the average values of \(u_{\text{CoCoA}}\) (Q4) and \(u_{\text{ret}}\) (Q5). To ensure this confidence is meaningful, we assess whether uncertainty estimates align with summary quality, measured by the proportion of correct claims.

Follow prior work \cite{vashurin2025uncertainty}, we quantify this alignment via the Prediction Rejection Ratio (PRR):
\[
\text{PRR} = \frac{\text{AUC}_{\text{unc}} - \text{AUC}_{\text{rnd}}}{\text{AUC}_{\text{oracle}} - \text{AUC}_{\text{rnd}}},
\]
where \(\text{AUC}_{\text{unc}}\) is obtained via uncertainty-based rejection, \(\text{AUC}_{\text{rnd}}\) is a random baseline, and \(\text{AUC}_{\text{oracle}}\) is an ideal oracle. Higher PRR values reflect better alignment between uncertainty and factual accuracy.

\subsection{Baselines}
We conduct a comparative analysis of (i) a LangChain SQL agent \footnote{\url{https://python.langchain.com/docs/integrations/tools/sql_database/}},  augmented with Python-based tools and leveraging the OpenAI-o4-mini model as its backbone, which executes database queries and code and produces one-shot summaries without uncertainty modeling (ii) our agent before GRPO training; (iii) our model after GRPO training, which incorporates uncertainty-aware reward shaping; and (iv) our GRPO-trained agent with post-output filtering as described in section \ref{sec:inference}.

\subsection{Results}

Our uncertainty-aware agent advances multi-table summarization, delivering significant improvements in summary quality and reliability across both test datasets, as evidenced in Tables \ref{tab:cancer} and \ref{tab:internal}. As the first to tackle this task with the MLOmics dataset, our approach sets a new benchmark, producing more claims with substantially higher correctness and usefulness ratios. Correctness increased from 1.5 to 9.9 average correct claims per summary in the cancer multi-omics dataset and from 0.9 to 8.4 in the internal dataset, a clear demonstration of the power of uncertainty-based rewards in curbing spurious outputs. Usefulness ratios rose from 0.60 to 0.78  on the internal dataset, reflecting enhanced schema navigation and evidence synthesis across diverse environments. 

These gains generalize across a proprietary schema-rich multi-omics corpus and the MLOmics benchmark, underscoring the agent’s adaptability. While the lack of prior work on this specific task/dataset combination highlights the pioneering nature of our results, they also outstrip the LangChain SQL-agent baseline (e.g., 3.6 vs. 9.9 correct claims in cancer, 3.0 vs. 8.4 internally), which lacks uncertainty modeling. Our approach sharpens uncertainty estimates, with retrieval entropy ($u_{\text{ret}}$) and summary uncertainty ($u_{\text{CoCoA}}$) decreasing, signaling more stable evidence acquisition and consistent outputs. The Prediction Rejection Ratio (PRR) improvements—rising to 0.45 (cancer) and 0.47 (internal) for CoCoA—validate that uncertainty signals serve as potent control mechanisms, aligning confidence with factual reliability and enhancing trustworthiness.

Focusing on the filtering step during inference, this component improved performance metrics, boosting correctness from 0.82 to 0.94 (cancer) and from 0.84 to 0.90 (internal), while usefulness climbed from 0.39 to 0.43 and 0.71 to 0.78, respectively. This underscores the critical role of inference-time refinement in producing reliable summaries across heterogeneous settings.

\begin{table}[t]
\centering
\caption{\textbf{Cancer Multi-Omics dataset performance.}
Average claims (Q1), correct claims (Q2), and useful claims (Q3) per
summary, with correctness/usefulness ratios. We also report uncertainty metrics
$u_{\text{CoCoA}}$ (Q4) and $u_{\text{ret}}$ (Q5); for each, the value outside parentheses is the uncertainty (↓),
and the value in parentheses is PRR (↑). Arrows in headers indicate the direction of better results.
The LangChain agent does not produce uncertainty metrics (shown as --).}
\vspace{0.5em}
\begin{adjustbox}{width=\textwidth}
\begin{tabular}{lccccc}
\toprule
\textbf{System} &
\textbf{\# Claims / summary $\uparrow$} &
\textbf{\# Correct / summary (ratio) $\uparrow$} &
\textbf{\# Useful / summary (ratio) $\uparrow$} &
$u_{\text{CoCoA}}$ $\downarrow$ \textbf{(PRR $\uparrow$)} &
$u_{\text{ret}}$ $\downarrow$ \textbf{(PRR $\uparrow$)} \\
\midrule
LangChain Agent & $5.4 \pm 0.7$ & $3.6 \pm 0.6$ ($0.67 \pm 0.04$) & $2.0 \pm 0.5$ ($0.37 \pm 0.03$) & -- & -- \\
Ours (Before Training) & $2.4 \pm 0.5$ & $1.5 \pm 0.4$ ($0.63 \pm 0.03$) & $0.9 \pm 0.3$ ($0.40 \pm 0.04$) & $0.47 \pm 0.05$ ($0.37 \pm 0.09$) & $0.84 \pm 0.06$ ($0.24 \pm 0.07$) \\
Ours ($R_{adapt}$, before filtering) & $10.2 \pm 1.3$ & $8.4 \pm 1.1$ ($0.82 \pm 0.03$) & $4.0 \pm 0.8$ ($0.39 \pm 0.04$) & $0.25 \pm 0.04$ ($0.38 \pm 0.09$) & $0.67 \pm 0.05$ ($0.25 \pm 0.06$) \\
Ours ($R_{adapt}$, after filtering) & $10.5 \pm 1.5$ & $9.9 \pm 1.2$ ($0.94 \pm 0.02$) & $4.5 \pm 0.9$ ($0.43 \pm 0.03$) & $0.19 \pm 0.03$ ($0.45 \pm 0.08$) & $0.44 \pm 0.04$ ($0.28 \pm 0.08$) \\
\bottomrule
\end{tabular}
\end{adjustbox}
\label{tab:cancer}
\end{table}

\begin{table}[t]
\centering
\caption{\textbf{Internal dataset performance.} Average claims (Q1), correct claims (Q2), and useful claims (Q3) per
summary, with correctness/usefulness ratios. We also report uncertainty metrics
$u_{\text{CoCoA}}$ (Q4) and $u_{\text{ret}}$ (Q5); for each, the value outside parentheses is the uncertainty (↓),
and the value in parentheses is PRR (↑). Arrows in headers indicate the direction of better results.
The LangChain agent does not produce uncertainty metrics (shown as --).}
\vspace{0.5em}
\begin{adjustbox}{width=\textwidth}
\begin{tabular}{lccccc} 
\toprule
\textbf{System} &
\textbf{\# Claims / summary} $\uparrow$ &
\textbf{\# Correct / summary (ratio)} $\uparrow$ &
\textbf{\# Useful / summary (ratio)} $\uparrow$ &
$\boldsymbol{u_{\text{CoCoA}}}$ $\downarrow$ \textbf{(PRR $\uparrow$)} &
$\boldsymbol{u_{\text{ret}}}$ $\downarrow$ \textbf{(PRR $\uparrow$)} \\
\midrule
LangChain Agent &
$4.5 \pm 0.8$ &
$3.0 \pm 0.6$ ($0.67 \pm 0.05$) &
$3.0 \pm 0.5$ ($0.65 \pm 0.04$) &
-- & -- \\
Ours (Before Training) &
$1.5 \pm 0.3$ &
$0.9 \pm 0.2$ ($0.60 \pm 0.03$) &
$0.9 \pm 0.2$ ($0.60 \pm 0.02$) &
$0.45 \pm 0.05$ ($0.39 \pm 0.09$) &
$0.84 \pm 0.04$ ($0.29 \pm 0.07$) \\
Ours ($R_{\text{adapt}}$, before filtering) &
$9.3 \pm 1.2$ &
$7.2 \pm 1.0$ ($0.84 \pm 0.03$) &
$6.6 \pm 0.7$ ($0.71 \pm 0.04$) &
$0.27 \pm 0.04$ ($0.42 \pm 0.08$) &
$0.65 \pm 0.07$ ($0.33 \pm 0.07$) \\
Ours ($R_{\text{adapt}}$, after filtering) &
$9.3 \pm 1.1$ &
$8.4 \pm 0.9$ ($0.90 \pm 0.02$) &
$7.2 \pm 0.8$ ($0.78 \pm 0.03$) &
$0.20 \pm 0.04$ ($0.47 \pm 0.08$) &
$0.42 \pm 0.06$ ($0.38 \pm 0.08$) \\
\bottomrule
\end{tabular}
\end{adjustbox}
\label{tab:internal}
\end{table}

\subsection{Ablation}
We study four factors that could explain our improvements: reward schedules, uncertainty signals, judge dependence, and inference thresholds. Tables and details can be found in Appendix \ref{app:ablat}.

\paragraph{Reward schedules.}

Reward shaping substantially affects optimization trajectories. Table~\ref{tab:schedules} compares $R_{zero}$, $R_{base}$, $R_{phase}$, $R_{step}$, and $R_{adapt}$. $R_{adapt}$ yields the highest useful-claims ratio (0.78 vs 0.30 for $R_{base}$ on Internal) and stronger PRR alignment. Learning curves (Fig.~\ref{fig:training_plots}) show $R_{adapt}$ avoids early collapse seen in $R_{base}$, indicating that adaptive weighting of uncertainty stabilizes training.

\paragraph{Uncertainty signals.}
We compared perplexity, CoCoA, entropy, and retrieval variance as reward signals within the $R_{judge}$ schedule. Perplexity yields a baseline Useful Ratio of $0.78$ with PRR 0.47. CoCoA improves calibration slightly (Useful Ratio $0.72$, PRR 0.50) but requires 2.6$\times$ more compute and adapts more slowly. Entropy ($0.76$, PRR 0.46) and retrieval variance ($0.76$, PRR 0.39) achieve stronger cost–benefit tradeoffs. The mechanism is straightforward: training purely for consistency encourages rigidity, while lighter signals adapt more flexibly. Full results are in Appendix Table~\ref{tab:signals}.

\paragraph{Judge robustness.}
Optimizing a single judge invites reward hacking \citep{ziegler2019fine, gao2023scaling}. We compared $R_{adapt}$ models scored by (i) our strong $R_{judge}$, (ii) a weaker LLM judge (GPT-4.1 Nano and Gemini 2.5 Flash Lite), and (iii) a 40-query human holdout. Correlations were moderate-to-strong ($r=0.62\pm0.08$ vs weak; $r=0.64\pm0.07$ vs human). Importantly, \emph{system rankings were identical}: $R_{adapt}$ $>$ $R_{step}$> $R_{phase}$ $>$ $R_{base}$. This indicates the gains are not artifacts of one evaluator.

\paragraph{Inference thresholds.}
Finally, we varied uncertainty thresholds $\kappa \in \{0.2,0.5,0.8\}$ for post-hoc filtering. Table~\ref{tab:inference} shows the trade-off: higher $\kappa$ reduces coverage but improves precision. $R_{adapt}$ models dominate at all thresholds.

\subsection{Prediction Results}
\label{sec:results_prediction}
Beyond evaluating summary correctness and usefulness, it is important to test whether the agent’s knowledge transfer produces meaningful downstream outcomes. Survival prediction provides such a test, connecting textual reasoning with a clinically relevant endpoint.
To perform this task, we prompt the agent to estimate survival times for held-out patients by leveraging in-context knowledge from summaries related to survival, rather than task-specific supervision. The predictions are evaluated using the concordance index (C-index), which measures how well predicted survival times align with ground-truth outcomes.
\begin{table}[t]
\centering
\caption{Concordance index (C-index) scores on the held-out test set. Results compare a LangChain baseline with our method under different refinement strategies ($R_{base}$, $R_{phase}$, $R_{step}$, $R_{adapt}$). Higher values indicate better predictive alignment.}
\vspace{0.5em}
\begin{adjustbox}{max width=\textwidth}
\begin{tabular}{lcccccc}
\toprule
\textbf{Model} & \textbf{LangChain Agent} & \textbf{Ours (Before Training)} & \textbf{Ours ($R_{base}$)} & \textbf{Ours ($R_{phase}$)} & \textbf{Ours ($R_{step}$)} & \textbf{Ours ($R_{adapt}$)} \\
\midrule
\textbf{C-Index} & 0.22 & 0.32 & 0.55 & 0.60 & 0.64 & 0.63 \\
\bottomrule
\end{tabular}
\end{adjustbox}
\label{tab:surv_table}
\end{table}
As shown in Table~\ref{tab:surv_table}, our framework consistently outperforms the LangChain baseline, with the largest improvement from $R_{step}$. The stable performance highlights the role of uncertainty-aware refinement in producing reliable predictions.

Additionally, it is important to note that untrained agents performed worse than random chance. They exhibited a tendency to systematically focus on incorrect features drawn from the literature, rather than accurately interpreting the dataset. This underscores the necessity of training and appropriate methodology to improve predictive performance.

\section{Discussion}
This work introduces uncertainty-aware LLM agents that explicitly incorporate retrieval and summary uncertainty into both training and inference, targeting the persistent challenge of reliable tabular summarization. Our main contribution is the shift from treating uncertainty as a post-hoc diagnostic to making it a 
\textbf{first-class control signal} that shapes optimization, guides agent behavior, and governs inference-time filtering.
Empirically, our framework achieves two notable outcomes. First, we observe consistent gains in factuality: uncertainty-aware agents produce nearly twice as many useful claims compared to baseline SQL agents, with improvements reflected in both automatic fact-checking and downstream survival analysis tasks. Second, uncertainty estimates themselves prove informative: the PRR roughly doubles, indicating that uncertainty correlates well with factual reliability. Together, these findings suggest that uncertainty-aware signals are not just auxiliary diagnostics but \textbf{actionable levers} for building more trustworthy agents.

Beyond raw performance, our study highlights an underexplored but critical design principle: \textbf{agents should know when not to answer}. By abstaining on high-uncertainty outputs and filtering synthetic data accordingly, our method aligns with a conservative, safety-first philosophy that is especially vital in biomedical applications. This aligns our work with a broader trend toward self-reflective and self-assessing agents, while providing concrete evidence that such mechanisms can enhance reliability in structured data environments.

While our experiments are conducted on biomedical multi-omics datasets, the framework is domain-agnostic and immediately applicable to other tabular contexts such as finance, e-commerce, or clinical EHR systems. Importantly, our design choices—entropy-based retrieval uncertainty, self-consistency signals, and GRPO-based training—are modular and can be integrated into existing table-agent pipelines without architectural overhaul.
Overall, the discussion we wish to emphasize is not that uncertainty eliminates hallucinations—indeed, some degree of error is inevitable—but that embedding uncertainty into the \textbf{decision loop} of an agent allows us to manage, calibrate, and ultimately trust these systems in ways post-hoc filtering cannot. We see this as a principled step toward agents that are transparent about their confidence and therefore safer for deployment in high-stakes settings.

\section{Limitations}

Our current evaluation is limited to biomedical multi-omics data. While this domain highlights the need for reliability in high-stakes settings, testing across finance, e-commerce, and other structured environments will demonstrate the broader generality of the framework.

We also rely on automated LLM-based judges for reward shaping and fact-checking. This enables scalable experimentation but could potentially induce bias. Expanding systematic human validation will be an important next step, and our uncertainty annotations can help guide such expert audits.

Finally, the method requires multiple rollouts (e.g., K=5) and CoCoA-based self-consistency, which add inference cost. Preliminary results suggest smaller K retains most benefits, and leveraging a lightweight uncertainty proxy eg. perplexity instead of CoCoA could make the approach more efficient.

\section{Conclusion}

This work shows that uncertainty can be treated not just as a diagnostic signal, but as an active control mechanism for agentic systems operating over structured data. By combining retrieval and summary uncertainty during both training and inference, our agent learns when to proceed and when to abstain, improving both correctness and safety in multi-table reasoning tasks. While early results suggest benefits for downstream analysis, open challenges remain in calibration, evaluation beyond proprietary datasets, and reducing inference costs. We see this as a step toward building agents that scale responsibly, and we invite the community to explore stronger uncertainty estimation methods, richer benchmarks, and ethical safeguards.

\section{Ethics}

\textbf{Compliance}: The main results of this paper are supported by publicly available datasets (MLOmics and other open cancer databases) under their original licenses. No patient-identifiable data was used. MLOmics data is de-identified and released under standard open science protocols.
Reproducibility: Code, prompts, and configuration files will be made available to support replication. Hyperparameters and training procedures are documented in Appendix \ref{app:Hyper}.

\textbf{Validation methodology}: Automated judge scores were validated against human expert assessment on a subset of outputs (N=40 queries) to ensure reliability. Agreement metrics and audit protocols are provided in supplementary materials.

\textbf{Limitations disclosure}: This system is designed for research exploration, not clinical decision-making. Expert oversight is required for any biomedical applications, and outputs should not be used as medical advice without appropriate validation.

\bibliographystyle{plain} 
\bibliography{bib} 

\appendix
\renewcommand{\thefigure}{A\arabic{figure}}
\renewcommand{\thetable}{A\arabic{table}}

\section{Additional Methods Details}
\label{app:details}

This section collects additional details about our setup that were omitted from the main text for clarity.



\paragraph{Summary Uncertainty.}

\textit{Perplexity.} For a summary token sequence $s_{1:T}$:  
\begin{equation}
u_{\text{Perp}}(s_{1:T}) = \exp\!\left(-\tfrac{1}{T}\sum_{t=1}^{T}\log p_\theta(s_t \mid s_{<t}, \text{context})\right), 
\label{eq:perp}
\end{equation}
where \( p_\theta(s_t \mid s_{<t}, \text{context}) \) represents the probability assigned by the model to token $s_t$ given the sequence of preceding tokens $s_{<t}$ and any task-specific contextual information. Lower perplexity implies higher model confidence in the token-level generation process.  
 
\textit{CoCoA.} \label{eq:CoCoA} We use the \textsc{CoCoA} metric \citep{vashurin2025uncertainty}, which enhances perplexity-based confidence -- relying solely on LLM probabilities and providing no information about the answer distribution -- with semantic self-consistency.

Given an actual output sequence $s^*$ and $K-1$ sampled sequences $s^{(k)}, k=1, ..., K-1$, we compute a consistency-based uncertainty metric \cite{vashurin2025uncertainty}
\begin{align*}
u_{\text{cons}}(s^*, \{ s^{(k)}\}) &= 1 - \frac{1}{K-1} \sum_{k=1}^{K-1} \text{sim}(s^{(k)}, s^{*}), \\
\end{align*}

where, for the similarity metric sim we use the RoBERTa-large cross-encoder model, fine-tuned on the
Semantic Textual Similarity benchmark dataset \cite{reimers2019sentence, liu2019roberta, cer2017semeval}.
Multiplying $u_{\text{cons}}(s^*,\{ s^{(k)}\})$ with the perplexity of $s^*$ produces the CoCoA metric $u_{\text{CoCoA}}(s^*,\{ s^{(k)}\} )$. 


\paragraph{Reward Design.}
\textit{Code Execution Reward.} To incentivize correct database interactions, the trajectory receives a reward based on the number of correctly executed SQL queries or Python code executions, with a stronger emphasis on rewarding initial successes to encourage learning.  Let \(x(\tau) \in \mathbb{N}\) be the number of correctly executed code actions in trajectory $\tau$. The code execution reward is:
\[
R_{code}(\tau) = \min\left(1, \frac{\log(10x(\tau)+1)}{\log(31)}\right),
\]
where the reward is capped at a maximum value of 1 for three correctly executed actions. This design aims to teach the model to produce correct executable code early in training. The reward cap ensures the model saturates the benefit from code execution once it reliably achieves three successful actions, encouraging it to focus on higher-level tasks, such as summary generation, as training progresses.

\textit{Exploration Judge Reward.} An external \texttt{o4-mini} LLM counts the number \(c(\tau)\) of grounded, non-overlapping atomic facts in the trajectory \(\tau\) that are relevant to the user's topic. The reward is:  
\[
R_{\text{Judge}}(\tau) = \min\left(\frac{c(\tau)}{20}, 1\right),
\]
promoting thorough database exploration to uncover relevant and diverse information. The normalization factor \(20\) reflects our empirical observation that trajectories with around 20 grounded, non-overlapping facts typically provide sufficient diversity and coverage for most queries. 
    
\textit{Summary Confidence Reward.} The inverse perplexity of the generated summary \(s(\tau)\) corresponding to the trajectory \(\tau\), serves as a measure of token-level confidence, normalized to \((0,1]\):  
\[
R_{\text{conf}}(\tau) = \frac{1}{u_{\text{Perp}}(s(\tau))}.
\]
While \(R_{\text{Judge}}\) promotes database exploration, \(R_{\text{conf}}\) incentivizes exploitation by rewarding low-uncertainty summaries. Consistency-based uncertainty metrics, such as CoCoA, are omitted during training to sidestep the high computational overhead of sampling.

\paragraph{Reward Schedules.}
 We explore various reward schedules over the 100 training steps \(t\) to balance exploration and exploitation. A summary of these schedules is provided in Table~\ref{tab:reward-schedules}. Constants are empirically chosen to balance the contributions of individual reward components, ensuring effective training dynamics. Ablation studies of these constants are left for future work.

\begin{table}[h!]
    \centering
    \caption{Summary of reward schedules, their formulas, and descriptions.}
    \small
    \label{tab:reward-schedules}
    \small
    \begin{adjustbox}{width=\textwidth}
    \begin{tabular}{|p{1.5 cm}|c|p{3cm}|} 
        \hline
        \textbf{Schedule} & \textbf{Formula} & \textbf{Description} \\ \hline
        Zero &
        \( R_{zero}(\tau) = R_{code}(\tau) + 4 R_{Judge}(\tau) \) &
        Does not use the uncertainty signal in reward. \\ \hline
        Baseline &
        \( R_{base}(\tau) = R_{code}(\tau) + 4 R_{Judge}(\tau) + \frac{1}{3} R_{\text{conf}}(\tau) \) &
        Uses a fixed combination of all three reward components. \\ \hline
        Two-Phase &
        \(\begin{aligned}
            R_{phase}(\tau) =
            \begin{cases}
                R_{code}(\tau) + 4 R_{Judge}(\tau), & \text{if } t \leq 50, \\
                R_{code}(\tau) + 4 R_{Judge}(\tau) + \frac{1}{3} R_{\text{conf}}(\tau), & \text{if } t > 50.
            \end{cases}
        \end{aligned} \) &
        Focuses on exploration during the first half, incorporates exploitation in the second training half. \\ \hline
        Stepwise  Addition &
        \(\begin{aligned}
            R_{step}(\tau) =
            \begin{cases}
                R_{code}(\tau) + 4 R_{Judge}(\tau), & \text{if } t \text{ mod } 10 \neq 0, \\
                 R_{code}(\tau) + 4 R_{Judge}(\tau) + 2 R_{\text{conf}}(\tau), & \text{if } t \text{ mod } 10 = 0.
            \end{cases}
        \end{aligned} \) &
        Periodically emphasizes exploitation every 10 steps. \\ \hline
        Adaptive Exploitation  &
        \(\begin{aligned}
            \alpha &= \exp\big(-50 \big(R_{\text{Judge}}(\tau) - \frac{1}{2}\big)^2\big), \\
            R_{adapt}(\tau) &=  R_{code}(\tau) + 4 R_{Judge}(\tau) + 2 \alpha R_{\text{conf}}(\tau).
        \end{aligned} \) &
        Initially promotes exploration, then gradually integrates exploitation, and tapers off to prevent generic summaries. \\ \hline
    \end{tabular}
    \end{adjustbox}
\end{table}

\paragraph{Episode algorithms.}

Figure \ref{alg:epi} describes algorithms for full training and inference episodes.

\begin{figure}[t]
\begin{algorithm}[H]
\caption{ Episode Algorithms for Training and Inference}  \begin{algorithmic}
\small
    \Procedure{SINGLE\_EPISODE}{$q, \mathcal{D}, \pi_\theta$, M}
        \State Initialize empty trajectory \(\tau \gets \{\}\)
        \For{$t = 1 ,...,$ M}
            \State Sample action \(a_t \sim \pi_\theta(\cdot \mid x_t)\)
            \If{$a_t$ is \texttt{SQLExecutor(query)}}
                \State \(r_t \gets\) Execute SQL query on \(\mathcal{D}\)
                \State Append \((a_t, r_t)\) to \(\tau\)
            \ElsIf{$a_t$ is \texttt{PythonTool(code)}}
                \State \(r_t \gets\) Execute Python code on relevant database parts
                \State Append \((a_t, r_t)\) to \(\tau\)
            \ElsIf{$a_t$ is \texttt{Schema(Table)}}
                \State \(r_t \gets\) Retrieve schema of the specified table
                \State Append \((a_t, r_t)\) to \(\tau\)
            \ElsIf{$a_t$ is \texttt{CommitSummary(summary)}}
                \State Extract summary \(s\), including token logits
                \State Append \((a_t, s)\) to \(\tau\)
                \State \textbf{break}
            \EndIf
        \EndFor
        \State \Return \(\tau, s\)
    \EndProcedure

    \Procedure{TRAINING\_EPISODE}{$q, \mathcal{D}, \pi_\theta$, M}
        \State \(\tau, s \gets \text{SINGLE\_EPISODE}(q, \mathcal{D}, \pi_\theta, M) \)
        \State Compute token-level perplexity \(u_{PPL}(s)\) 
        \State Compute rewards \(R_{Judge} (\tau), R_{code}(\tau), R_{conf}(\tau)\) 
        \State Combine \(R_{Judge} (\tau), R_{code}(\tau), R_{conf}(\tau)\) to compute terminal reward \(R(\tau)\)
        \State Store \((\tau, R(\tau))\) for GRPO update
    \EndProcedure

    \Procedure{INFERENCE\_EPISODE}{$q, \mathcal{D}, \pi_\theta$, K}
        \State Initialize \(\mathcal{S} \gets \{\}\) and \(\mathcal{T} \gets \{\}\)
        \For{$k = 1 ,..., K$}
            \State \(\tau_k, s_k \gets \text{SINGLE\_EPISODE}(q, \mathcal{D}, \pi_\theta, M)\)
            \State Append \(s_k\) to \(\mathcal{S}\), \(\tau_k\) to \(\mathcal{T}\)
        \EndFor
        \State Compute summary uncertainty \(u_{\text{CoCoA}}(\mathcal{S})\)
        \State Compute retrieval uncertainty \(u_{\mathrm{ret}}(\mathcal{T})\) from SQL queries in trajectories
        \State $\Tilde{\tau}, \Tilde{s} \gets \text{(trajectory, summary) pair with lowest-perplexity summary from zip}(\mathcal{T}, \mathcal{S}) $
        \State Store \((\Tilde{\tau}, \Tilde{s}, u_{\text{CoCoA}}(\mathcal{S}), u_{\mathrm{ret}}(\mathcal{T}))\)
    \EndProcedure
\end{algorithmic}
\end{algorithm}
\caption{Episode algorithms. Training uses a single episode to compute terminal reward based on code execution, confidence and exploration. Inference samples multiple episodes to compute summary and retrieval uncertainties.}
\label{alg:epi}
\end{figure}
\section{Datasets}
\label{app:datasest}

This section describes the datasets used in our experiments.

\subsection{Internal Multi-Omics Dataset}
The internal dataset is built from multi-layered omics data. While the specific table contents cannot be disclosed, its structure can be summarized as:
\begin{itemize}
  \item \textbf{Architecture:} Multi-layered, with each layer corresponding to a distinct omics modality (e.g., transcriptomics, proteomics, metabolomics).
  \item \textbf{Scale:} Each layer consists of between tens and hundreds of relational tables.
  \item \textbf{Topologies:}
  Two primary schema structures are observed:
  (a) a \emph{tree-like hierarchy}, in which child tables branch recursively from root entities, and
  (b) a \emph{broad schema}, in which many tables connect directly to a central entity.
\end{itemize}
These schema variations provide structurally distinct environments that stress-test an agent’s ability to adapt to different database organizations.

\subsection{Dataset Schematic for Internal Multi-Omics Dataset}
\begin{figure}[ht]
  \centering
    \includegraphics[width=\linewidth]{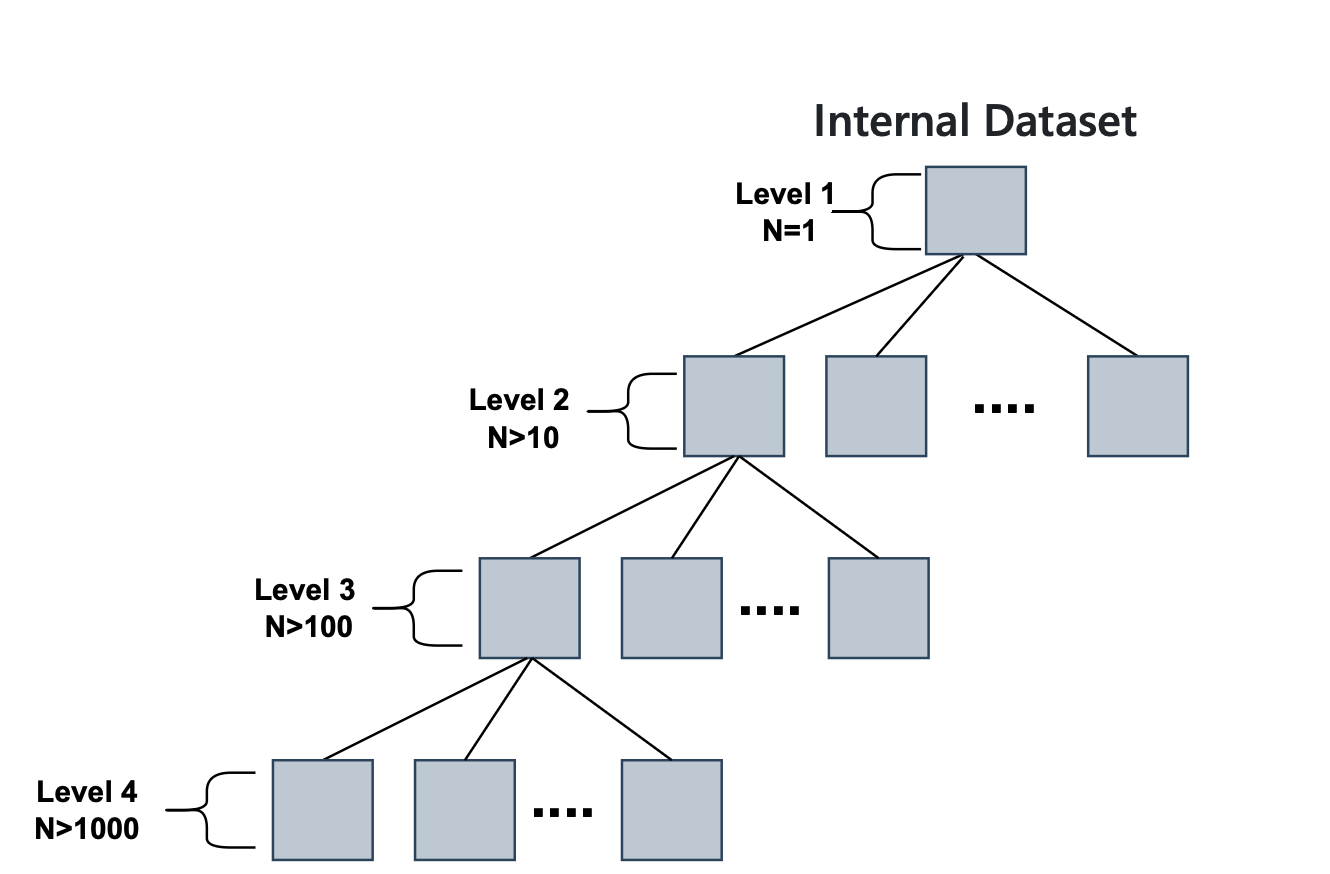}
  \vspace{1em}
  \caption{Internal multi-omics dataset showing tree-like schema topology.}
  \label{fig:internal_dataset_schematic}
\end{figure}

Figure \ref{fig:internal_dataset_schematic} contains dataset Schematic for the Internal Multi-Omics Dataset.

\subsection{MLOmics: Cancer Multi-Omics Database for Machine Learning}
MLOmics~\cite{yang2025mlomics} is an open multi-omics dataset comprising 8,314 patients across 32 cancer types. It provides four standardized omics modalities:
\begin{itemize}
  \item \textbf{mRNA expression:} Gene-level transcriptional profiles.
  \item \textbf{microRNA expression:} Small noncoding RNAs regulating gene expression.
  \item \textbf{DNA methylation:} CpG site methylation fractions representing epigenetic regulation.
  \item \textbf{Copy number variation (CNV):} Segment-level gene copy alterations.
\end{itemize}
Each modality is released in three feature versions:
\begin{itemize}
  \item \emph{Original}: full feature set,
  \item \emph{Aligned}: subsets harmonized across modalities,
  \item \emph{Top}: statistically filtered subsets (ANOVA-based).
\end{itemize}
MLOmics additionally includes baseline machine learning benchmarks (6–10 methods), clustering and survival analyses, and external knowledge integration (STRING, KEGG). These resources make it a reproducible benchmark for developing and evaluating uncertainty-aware agents.

\subsection{Dataset Schematic for MLOmics}
\begin{figure}[ht]
  \centering
    \includegraphics[width=\linewidth]{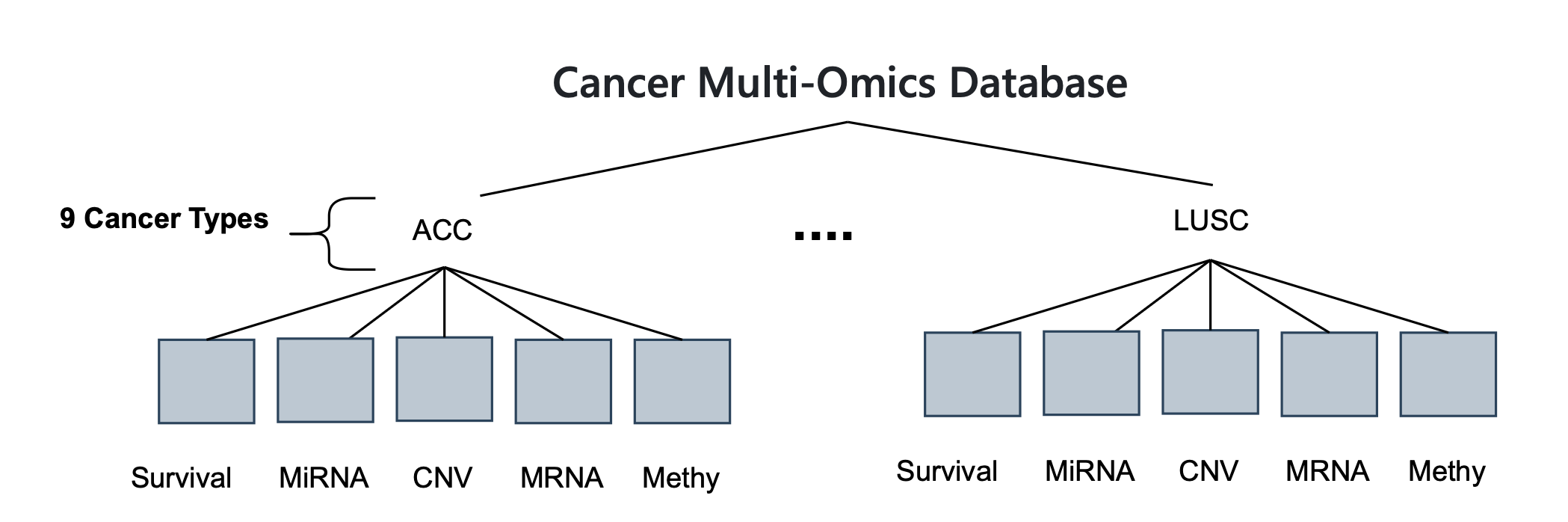}
  \caption{Public MLOmics dataset, with standardized parallel modalities spanning 9 cancer types.}
  \label{fig:mlomics_dataset_schematic}
\end{figure}

Figure \ref{fig:mlomics_dataset_schematic} contains dataset Schematic for Cancer MLOmics Dataset.
\section{Summary Tasks} \label{app:summ}

\label{app:task_templates}

This section provides examples task templates used in training and inference for the Cancer MLOmics Dataset. Each task outlines specific objectives and details the steps required to obtain relevant information about different cancer types using molecular data. The complete list of tasks will be released on GitHub upon completion.

\begin{tcolorbox}[colback=yellow!10!white, colframe=yellow!75!black, title=Task 1: Basic Cancer-Survival Characterization]
\textbf{Objective:} For a specified cancer type \texttt{CANCER\_TYPE}, answer the following questions:
\begin{enumerate}
    \item How many patients are in the training set?
    \item What is the median survival time?
    \item What is the event rate (percentage of deaths)?
    \item Describe the survival distribution.
    \item Compare this cancer's survival patterns to other cancers in the database. 
\end{enumerate}
\end{tcolorbox}

\vspace{10pt} 

\begin{tcolorbox}[colback=blue!10!white, colframe=blue!75!black, title=Task 2: Molecular Data Profile]
\textbf{Objective:} For a specified cancer type \texttt{CANCER\_TYPE}, analyze each omic layer:
\begin{enumerate}
    \item Data distribution characteristics for each omic type.
    \item Missing value analysis. 
    \item Create a molecular profile summary specific to this cancer type.
\end{enumerate}
\end{tcolorbox}

\vspace{10pt}

\begin{tcolorbox}[colback=green!10!white, colframe=green!75!black, title=Task 3: Cancer-Specific Biomarkers]
\textbf{Objective:} For a specified cancer type \texttt{CANCER\_TYPE}, identify and analyze biomarkers:
\begin{enumerate}
    \item Identify top survival-associated features from each omic type:
    \begin{itemize}
        \item Top 20 mRNA features
        \item Top 20 miRNA features
        \item Top 20 methylation sites
        \item Top 20 CNV regions
    \end{itemize}
    \item Analyze their biological relevance.
    \item Compare with known markers for this cancer type.
    \item Create a prioritized biomarker list.
\end{enumerate}
\end{tcolorbox}

\vspace{10pt}

\begin{tcolorbox}[colback=orange!10!white, colframe=orange!75!black, title=Task 4: Multi-omic Integration]
\textbf{Objective:} For a specified cancer type \texttt{CANCER\_TYPE}, integrate various omic layers:
\begin{enumerate}
    \item Find correlations between features across different omic types.
    \item Identify multi-omic patterns associated with survival.
    \item Create an integrated molecular profile.
    \item Highlight unique molecular characteristics of this cancer type.
\end{enumerate}
\end{tcolorbox}

\vspace{10pt}

\begin{tcolorbox}[colback=purple!10!white, colframe=purple!75!black, title=Task 5: Clinical-Molecular Summary]
\textbf{Objective:} Create a comprehensive summary for a specified cancer type \texttt{CANCER\_TYPE}:
\begin{enumerate}
    \item Key survival characteristics.
    \item Most important molecular features.
    \item Multi-omic patterns.
    \item Clinical-molecular associations.
    \item Comparison with other cancer types.
    \item Potential clinical implications.
\end{enumerate}
\end{tcolorbox}


\section{Hyperparameters} \label{app:Hyper}

The backbone of the model used in this work is Qwen2.5-14B-Instruct, implemented within the ART framework. Training was conducted on 1$\times$NVIDIA A100 80GB GPU, with a total computational cost of approximately \texttt{22} GPU-hours per model. We use sampling defaults of $M = K = 5$.

The training process employs Grouped Relative Policy Optimization (GRPO) to optimize the summarization agent. We set the clipping parameter $\epsilon = 0.2$ and the KL penalty weight $\beta = 0.01$. The learning rate is defined as \texttt{5e-5}, selected after searching for optimal values in the range between \texttt{1e-7} and \texttt{1e-4}. The model is allowed up to \texttt{6} tool calls per query for performing retrievals and summary generation, determined through a search over \texttt{4–10} tool calls per query, where only marginal improvements were observed beyond \texttt{6} tool calls. 

Training is conducted in mini-batches consisting of \texttt{3 groups per step}, with each group containing \texttt{4 rollouts}, ensuring that every query is processed multiple times as part of GRPO optimization. Each training run spans \texttt{4 epochs}.

All code, prompts, and configuration files will be released to ensure reproducibility.

\section{Ablation Details}
\label{app:ablat}

\subsection{Reward schedules}
We evaluate five reward schedules ($R_{\text{zero}}$, $R_{\text{base}}$, $R_{\text{phase}}$, $R_{\text{step}}$, and $R_{\text{adapt}}$ with definitions in Table \ref{tab:reward-schedules}) to analyze the impact of uncertainty during training (Table~\ref{tab:schedules}). The $R_{\text{zero}}$ schedule, which excludes uncertainty rewards, has the worst correct claims ratio of 0.27 due to frequent hallucinated claims with high uncertainty. 

\begin{figure}[ht]
    \centering
    \includegraphics[width=\textwidth]{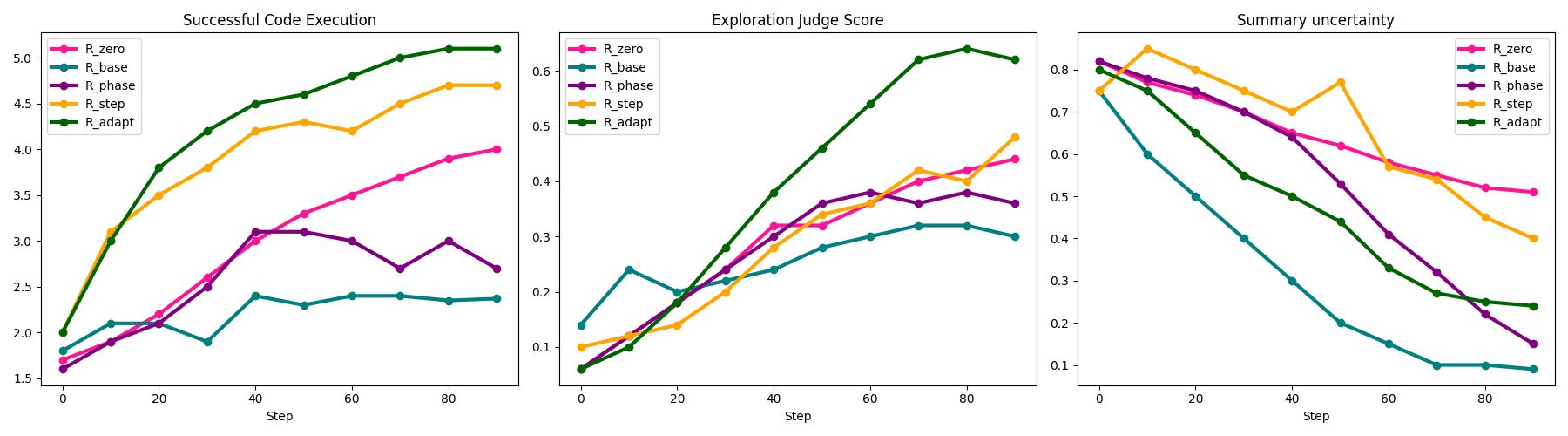}
    \caption{Performance metrics ($R_{\text{Code}}$, $R_{\text{Judge}}$, and uncertainty) during 100 training steps under different reward schedules ($R_{\text{zero}}, R_{\text{base}}, R_{\text{phase}}, R_{\text{step}}, R_{\text{adapt}}$).}
    \label{fig:training_plots}
\end{figure}

$R_{\text{base}}$, applying uncertainty rewards from the start, improves the correct claims ratio to 0.64 but achieves limited exploration (see $R_{code}$ and $R_{Judge}$ in Figure \ref{fig:training_plots}), leading to shallow summaries with useful claims ratios of 0.30 for the Internal dataset and 0.25 for Cancer Multi-Omics. 

To address these limitations, $R_{\text{phase}}$ defers uncertainty rewards to encourage early exploration. It raises the correct claims ratio to 0.67 and improves useful claims ratios to 0.50 on Internal and 0.41 on Cancer Multi-Omics, though outputs remain conservative and shallow due to excessive uncertainty minimization,  as reflected by summary uncertainty trends in Figure~\ref{fig:training_plots}.

$R_{\text{step}}$ introduces rewards periodically, boosting useful claims ratios to 0.55 (Internal) and 0.44 (Cancer Multi-Omics). However, abrupt uncertainty application every tenth step causes instability, reflected in unsmooth training plots in Figure~\ref{fig:training_plots} and inconsistent PRR values such as 0.32 for $u_{\text{CoCoA}}$ on Internal. 

Finally, $R_{\text{adapt}}$ dynamically adjusts uncertainty rewards, integrating them smoothly throughout training. This yields the best performance: correct claims ratios of 0.90 and 0.94, and useful claims ratios of 0.78 (Internal) and 0.43 (Cancer Multi-Omics), with strong uncertainty alignment (e.g., PRR of 0.47 for $u_{\text{CoCoA}}$).

\begin{table}[h]
\centering
\caption{Reward schedule ablations on both the Internal and the Multi-Omics Cancer Dataset. Average number of claims per summary, claim correctness and usefulness ratio, along with uncertainty metrics $u_{\text{CoCoA}}$ and $u_{ret}$ with PRR scores indicating alignment with correctness.}
\vspace{0.5em}
\begin{adjustbox}{max width=\textwidth}
\begin{tabular}{lccccccccccc}
\toprule
\textbf{Schedule} & \multicolumn{3}{c}{\textbf{Internal}} & \multicolumn{2}{c}{\textbf{Internal (UQ)}} & \multicolumn{3}{c}{\textbf{Cancer}} & \multicolumn{2}{c}{\textbf{Cancer (UQ)}} \\
\cmidrule(lr){2-4} \cmidrule(lr){5-6} \cmidrule(lr){7-9} \cmidrule(lr){10-11}
& \textbf{\# Claims} & \textbf{Correct Ratio} & \textbf{Useful Ratio} & $u_{\text{CoCoA}}$/PRR & $u_{ret}$/PRR & \textbf{\# Claims} & \textbf{Correct Ratio} & \textbf{Useful Ratio} & $u_{\text{CoCoA}}$/PRR & $u_{ret}$/PRR \\
\midrule
$R_{zero}$  & 5.2 $\pm$0.3 & 0.27 $\pm$0.05 & 0.27 $\pm$0.02 & 0.51/0.35 & 0.86/0.25 & 5.5$\pm$0.4 & 0.33 $\pm$0.06 & 0.29 $\pm$0.03 & 0.49/0.36 & 0.87/0.24 \\
$R_{base}$  & 4.5 $\pm$0.4 & 0.64 $\pm$0.04 & 0.30 $\pm$0.03 & { 0.13/0.33} & 0.72/0.28 & 5.2 $\pm$0.5 & 0.65 $\pm$0.04 & 0.25 $\pm$0.03 & {0.14/0.31} & 0.75/0.26 \\
$R_{phase}$ & 6.0 $\pm$0.5 & 0.67 $\pm$0.03 & 0.50 $\pm$0.03 & 0.15/0.39 & 0.65/0.33 & 6.8 $\pm$0.6 & 0.66 $\pm$0.03 & 0.41 $\pm$0.03 & 0.17/0.34 & 0.68/0.28 \\
$R_{step}$  & 8.3 $\pm$0.6 & 0.75 $\pm$0.03 & 0.55 $\pm$0.02 & 0.22/0.32 & 0.58/0.32 & 9.0 $\pm$0.7 & 0.78 $\pm$0.02 & \textbf{0.44 $\pm$0.03} & 0.25/0.39 & 0.61/0.29 \\
$R_{adapt}$ & 9.3 $\pm$1.1 & \textbf{0.90 $\pm$0.02} & \textbf{0.78 $\pm$0.03} & {0.20/0.47} & {0.42/0.38} & 10.5 $\pm$1.5 & \textbf{0.94 $\pm$0.01} & {0.43 $\pm$0.03} & {0.19/0.45} & {0.44/0.28} \\

\bottomrule
\end{tabular}
\end{adjustbox}
\label{tab:schedules}
\end{table}

\subsection{Uncertainty signals}
Table~\ref{tab:signals} compares four uncertainty reward signals. CoCoA improves consistency but is compute-inefficient.
\begin{table}[h]
\centering
\small
\begin{tabular}{lccc}
\toprule
\textbf{Signal} & \textbf{Useful Ratio} & \textbf{PRR} & \textbf{Relative Cost} \\
\midrule
Perplexity & 0.78 $\pm$0.03 & 0.47 & 1.0 \\
CoCoA & 0.72 $\pm$0.03 & 0.50 & 2.6 \\
Entropy & 0.76 $\pm$0.02 & 0.46 & 1.0 \\
Retrieval variance & 0.76 $\pm$0.02 & 0.39 & 2.1 \\
\bottomrule
\end{tabular}
\caption{Uncertainty signal ablations (internal dataset, $R_{adapt}$ schedule).}
\label{tab:signals}
\end{table}

\subsection{Judge robustness}
Table~\ref{tab:judges} gives correlations between $R_{judge}$, a weak LLM judge, and human labels. Preserved ranking: $R_{adapt}$ $>$ $R_{step}$> $R_{phase}$ $>$ $R_{base}$
\begin{table}[h]
\centering
\small
\begin{tabular}{lcc}
\toprule
\textbf{Comparison} & \textbf{Pearson $r$} & \textbf{Ranking preserved?} \\
\midrule
$R_{judge}$ vs Weak Judge & 0.62 $\pm$0.08 & Yes \\
$R_{judge}$ vs Human (40q) & 0.64 $\pm$0.07 & Yes \\
\bottomrule
\end{tabular}
\caption{Judge robustness. Rankings were consistent across reward schedules  (internal dataset).}
\label{tab:judges}
\end{table}


\subsection{Inference thresholds}
Table~\ref{tab:inference} shows coverage–accuracy tradeoffs for different thresholds $\kappa$. 
\begin{table}[h]
\centering
\small
\begin{tabular}{lcccc}
\toprule
\textbf{Method} & \textbf{Threshold $\kappa$} & \textbf{Coverage (\%)} & \textbf{Useful Ratio} & \textbf{PRR} \\
\midrule
$R_{adapt}$ & 0.5 & 70 & \textbf{0.78 $\pm$0.03} & \textbf{0.47} \\
$R_{adapt}$ & 0.2 & 95 & 0.72 $\pm$0.03 & 0.43 \\
$R_{adapt}$ & 0.8 & 40 & 0.85 $\pm$0.02 & 0.50 \\
\bottomrule
\end{tabular}
\caption{Inference thresholds. Post-hoc filtering improves slightly but underperforms full uncertainty-aware training (internal dataset).}
\label{tab:inference}
\end{table}

\end{document}